\documentclass[11pt,3p,authoryear]{elsarticle}

\journal{International Journal of Forecasting}
\biboptions{longnamesfirst}
\usepackage{amsmath, amsfonts, amssymb, amsthm}
\usepackage{xspace}
\usepackage{enumitem}
\usepackage{dsfont}	
\usepackage{url}
\usepackage{tikz}    
\usepackage{booktabs}
\usepackage{multirow}
\usepackage{worldflags}
\usepackage{subcaption}
\usetikzlibrary{arrows,positioning,shapes}
\usetikzlibrary{shapes.multipart}
\usetikzlibrary{arrows.meta}
\usetikzlibrary{decorations.pathreplacing}
\usepackage{rotating}
\usepackage{geometry}
\usepackage{hyperref}
\usepackage[table]{xcolor}
\definecolor{darkgreen}{rgb}{0.00,0.70,0.00}
\definecolor{darkblue}{rgb}{0.00,0.00,0.70}

\newcommand{\Ind}[1]    {\mathds{1}{\{#1\}}}

\newcommand{\V}{\ensuremath{\mathcal{V}}\xspace}

\newcommand{\bX}{\ensuremath{\mathbf X}\xspace}

\newcommand{\bA}{\ensuremath{\mathbf A}\xspace}

\newcommand{\bh}{\ensuremath{\mathbf h}\xspace}
\newcommand{\bW}{\ensuremath{\mathbf W}\xspace}

\begin{document}

\begin{frontmatter}

\title{Graph Neural Networks for Electricity Load Forecasting}

\author[cb,edf]{Eloi Campagne\corref{cor}}
\cortext[cor]{Corresponding author}
\ead{eloi.campagne@ens-paris-saclay.fr}

\author[edf]{Yvenn Amara-Ouali}
\author[edf]{Yannig Goude}
\author[cb,edf]{Itai Zehavi}
\author[cb]{Argyris Kalogeratos}

\address[cb]{Centre Borelli, Ecole Normale Supérieure Paris-Saclay, Gif-sur-Yvette, France}
\address[edf]{EDF Lab, Palaiseau, France}


\begin{abstract}
Forecasting electricity demand is increasingly challenging as energy systems become more decentralized and intertwined with renewable sources. Graph Neural Networks (GNNs) have recently emerged as a powerful paradigm to model spatial dependencies in load data while accommodating complex non-stationarities. This paper introduces a comprehensive framework that integrates graph-based forecasting with attention mechanisms and ensemble aggregation strategies to enhance both predictive accuracy and interpretability. Several GNN architectures---including Graph Convolutional Networks, GraphSAGE, APPNP, and Graph Attention Networks---are systematically evaluated on synthetic, regional (France), and fine-grained (UK) datasets. Empirical results demonstrate that graph-aware models consistently outperform conventional baselines such as Feed Forward Neural Networks and foundation models like TiREX. Furthermore, attention layers provide valuable insights into evolving spatial interactions driven by meteorological and seasonal dynamics. Ensemble aggregation, particularly through bottom-up expert combination, further improves robustness under heterogeneous data conditions. Overall, the study highlights the complementarity between structural modeling, interpretability, and robustness, and discusses the trade-offs between accuracy, model complexity, and transparency in graph-based electricity load forecasting.
\end{abstract}


\end{frontmatter}

\section{Introduction}\label{sec:intro}

The stability of power systems depends critically on maintaining an equilibrium between generation and demand, a task complicated by the fact that electricity cannot be stored efficiently and must therefore be produced in real time. While ongoing research explores ways to enhance the flexibility of various generation technologies, including nuclear energy \citep{beja2025enhancing}, accurate short-term demand forecasts remain indispensable for market participants and system operators. The transition toward decentralized grid architectures introduces novel sources of uncertainty, adding complexity to forecasting tasks. At the same time, the growing share of intermittent renewable resources, such as wind and solar power, amplifies variability at multiple spatial scales due to the heterogeneous distribution of production sites. External shocks, notably the Covid-19 crisis and subsequent economic downturn, have further disrupted traditional load patterns, underscoring the importance of methods capable of handling non-stationary behaviors \citep{alasali2021impact}. Fortunately, the increasing availability of geolocated and disaggregated load data provides new opportunities for models that can leverage such information to reduce forecast uncertainty \citep{obst2021adaptive, devilmarest2021statespace}. Parallel advances in adaptive forecasting techniques have also shown promise, demonstrating measurable improvements in predictive accuracy for aggregate electricity demand \citep{bregere2022online, antoniadis2022hierarchical}.
In this work, we present a unified framework that integrates graph-based forecasting, attention mechanisms, and ensemble aggregation strategies for electricity demand prediction. Our contributions are threefold:
\begin{itemize}
    \item we benchmark a broad range of GNN architectures---including GCN, GraphSAGE, APPNP, and attention-based variants---on synthetic, regional (France), and fine-grained (UK) datasets;
    \item we analyze the explanatory power of attention weights, highlighting evolving patterns of spatial interactions linked to external drivers;
    \item we show that ensemble aggregation consistently enhances robustness of forecasts, with learned bottom-up strategies systematically outperforming naive baselines.
\end{itemize}
This study thus bridges the gap between predictive performance, interpretability, and robustness, providing insights for the deployment of graph-based methods in operational forecasting. 

\paragraph{Reproducibility and Open Resources}
All experiments, datasets, and hyperparameter configurations used in this study are made openly available. 
The complete implementation, including data preprocessing scripts, graph construction pipelines, and training procedures, is hosted on 
\href{https://github.com/eloicampagne/GraphToolbox}{GitHub}. 
Processed datasets (French regional load and UK residential subsets), along with synthetic benchmarks, can be accessed via 
\href{https://zenodo.org/records/17453409?token=eyJhbGciOiJIUzUxMiJ9.eyJpZCI6IjJkMzEwODAzLWM2NDMtNDljZi05MTA1LTA3Mjk1NWQ5YmRhMCIsImRhdGEiOnt9LCJyYW5kb20iOiIzN2RjMjZiNDhiNzc5YzQ3NzQ5NTQ5ZDZmN2E0YTc3ZCJ9.WIfAN3nX9wRLfaAQim_BfQmiqiPkwKRKGr-lF2m2cOJM0d0O8rofN76WmcMCE6Pq-ushBaoIB2yZN_C3uKsxWQ}{Zenodo}. 
The codebase is built around the \emph{PyTorch Geometric} library \citep{fey2019fast,fey2025pyg}, which provides modular and efficient implementations of graph neural network layers and message passing schemes. 

\section{Related Work}\label{sec:related}

Forecasting electricity demand and renewable generation is a central task for energy system management, as accurate predictions guide decisions on market operations and grid optimization. Probabilistic approaches have been widely investigated in this context. Generalized Additive Models (GAMs) have long been effective for conditional mean forecasting \citep{fan2010forecast, fan2011short, pierrot2011short}, and extensions such as quantile GAMs \citep{fasiolo2021fast} or Generalized Additive Models for Location Scale and Shape (GAMLSS) \citep{gilbert2023probabilistic} allow the modeling of entire distributions. While these models perform well under stationary conditions, recent structural shifts---including the Covid-19 crisis \citep{jiang2021impacts} and the surge in European electricity prices \citep{doumeche2023human}---highlight their limitations in highly dynamic environments. To address this issue, adaptive extensions of GAMs combining online learning and filtering techniques have been proposed, improving robustness to evolving data regimes \citep{devilmarest2023adaptive}. Nevertheless, these approaches are primarily designed for aggregated demand, whereas the increasing decentralization of electricity networks calls for methods capable of handling spatially distributed signals \citep{williams2020electricity}.

Deep learning has proven highly versatile for energy applications, particularly when rich and heterogeneous datasets are available \citep{massaoudi2021deep}. Recurrent Neural Networks (RNNs), and in particular Long Short-Term Memory (LSTM) models, have achieved strong performance by capturing temporal dependencies in load series \citep{shi2017deep, marino2016building}. Yet, these models do not explicitly incorporate the structural relations between load nodes, which has motivated the exploration of Graph Neural Networks (GNNs). GNNs have demonstrated their ability to capture complex spatio-temporal dependencies in various domains, such as traffic forecasting \citep{guo2019attention, chen2020dynamic}, and have recently been applied to power systems. Notable contributions include hybrid GNN frameworks for wind power forecasting \citep{jiang2023buaa_bigscity} and more comprehensive graph-based models integrating external drivers \citep{jiang2024power}.

Building upon these developments, hybrid approaches have emerged to combine the interpretability of classical statistical models with the expressive power of deep learning. Neural Additive Models (NAMs) \citep{agarwal2021neural} extend the transparency of GAMs within a neural architecture, offering feature-wise decomposition while retaining non-linear flexibility. Their probabilistic variants further enable distributional forecasting in a data-driven framework \citep{thielmann2024neural}. These methods exemplify a broader effort to reconcile predictive accuracy and interpretability in complex forecasting systems.

In parallel, recent advances in GNN architectures explicitly address spatial reasoning and temporal dynamics through attention and multi-scale mechanisms. The GAT-LSTM model \citep{orji2025enhanced} integrates graph attention with recurrent units to jointly capture correlations between nodes and temporal evolution in load profiles, while embedding physical line attributes (e.g. capacities, losses) directly in the attention mechanism. The Multi-Resolution Graph Neural Network (MRGNN) \citep{mo2025mrgnn} leverages frequency-domain filtering and multi-scale aggregation to enhance robustness against noise and multi-timescale variability. Meanwhile, the Spatial-Temporal Dynamic Graph Transformer (SDGT) \citep{zhu2025short} introduces an adaptive Transformer capable of updating graph topology based on evolving temporal and semantic dependencies, emphasizing dynamic and structure-aware modeling.

Beyond graph-based architectures, the 
advent of foundation models has opened new perspectives for time-series forecasting. These large-scale pretrained architectures aim to generalize across heterogeneous temporal domains while retaining adaptability to local contexts. In the energy sector, PriceFM \citep{yu2025pricefm} is a spatio-temporal foundation model 
 designed for probabilistic electricity price forecasting in Europe. It embeds a graph-based inductive bias to represent market interconnections, and incorporates a distance-based graph decay to learn spatial dependencies across interconnected regions. More broadly, models such as TimesFM, TimeGPT-1, Aurora, and GridFM 
illustrate the transition toward unified, pretrained, and multimodal forecasting paradigms \citet{ferdaus2025foundation}. 

Overall, this body of research highlights a growing convergence between interpretable statistical modeling, graph-based learning, and foundation-level transfer architectures. The present work builds upon this evolution by systematically evaluating GNN architectures for electricity load forecasting, emphasizing interpretability, robustness, and reproducibility in an operational context.

\section{Preliminaries and Background} \label{sec:preliminaries}

We denote a graph by $\mathcal{G} = (\mathcal{V}, \mathcal{E})$, where $\mathcal{V}$ is the set of nodes and $\mathcal{E}$ the set of edges. An edge $(u,v) \in \mathcal{E}$ encodes a relationship or interaction between nodes $u$ and $v$. Graphs can be represented by a binary adjacency matrix $\bA \in \mathbb{R}^{N \times N}$, where $A_{uv}=1$ indicates the presence of an edge, or more generally by a weighted adjacency matrix $\bW$, where $W_{uv}$ quantifies the strength of the connection. The neighborhood of a node $v$ is written $\mathcal{N}_v = \{u \in \mathcal{V}\,\mid\, W_{uv} > 0\}$.
GNNs operate through message passing, an iterative process where nodes update their representations by aggregating information from their neighbors. At layer $\ell+1$, the hidden representation of node $v$, denoted $\bh_v^{(\ell+1)}$, is computed as
\begin{align*}
\bh_v^{(\ell+1)} = \mathsf{UPDATE}^{(\ell+1)}\!\Big(\bh_v^{(\ell)},~\mathsf{AGGREGATE}^{(\ell+1)}\!\big({\bh_u^{(\ell)} \,\mid\, u \in \mathcal N_v}\big)\Big),
\quad \bh_v^{(0)} = \bX_v,
\end{align*}
where $\mathsf{AGGREGATE}$ is typically a permutation-invariant operator (e.g. \texttt{sum}, \texttt{mean}, or \texttt{max}), and $\mathsf{UPDATE}$ applies a nonlinear transformation. By stacking $L$ layers, a node can incorporate information from its $L$-hop neighborhood, progressively expanding its receptive field.
This formulation can be viewed as a generalization of convolutional operations to irregular domains. The message passing mechanism acts as a graph convolution, extracting localized relational features while preserving the underlying topological structure of the data. As illustrated in Figure~\ref{fig: message_passing}, a message passing layer operates at three interconnected levels---node $V_{\ell}$, edge $E_{\ell}$, and global $U_{\ell}$ ---that are jointly updated through propagation and update functions. The propagation operators $\rho$ define how information is aggregated across the graph, for instance from nodes to edges $\rho_{V_\ell \to E_\ell}$, from edges to nodes $\rho_{E_\ell \to V_\ell}$, or from both to the global representation $\rho_{V_\ell \to U_\ell}$, $\rho_{E_\ell \to U_\ell}$. The update functions $\Phi = (\phi^v, \phi^e, \phi^u)$ then transform these aggregated messages into new latent representations for nodes, edges, and the global graph state, respectively $V_{\ell+1} = \phi^v(V_\ell, \rho_{E_\ell \to V_\ell}, U_\ell)$, $E_{\ell+1} = \phi^e(E_\ell, \rho_{V_\ell \to E_\ell}, U_\ell)$ and
$U_{\ell+1} = \phi^u(U_\ell, \rho_{V_\ell \to U_\ell}, \rho_{E_\ell \to U_\ell})$.
A classical example of this formulation is the Graph Convolutional Network (GCN), where the propagation operator aggregates neighboring node features through the normalized adjacency matrix $\tilde{\bA}$, and the update function is a linear transformation followed by a non-linearity:
\begin{align*}
    \bh_v^{(\ell+1)} = \sigma\!\left(\sum_{u \in \mathcal N_v \cup \{v\}} \tilde{A}_{uv}\,\bW^{(\ell)} \bh_u^{(\ell)}\right)
    = \sigma\!\left(\rho_{E_{\ell} \rightarrow V_{\ell}}\left(\bh^{(\ell)}\right)\,\bW^{(\ell)}\right),
\end{align*}
where $\rho_{E_{\ell} \rightarrow V_{\ell}}$ corresponds to the normalized neighborhood aggregation, and $\phi^v(\cdot) = \sigma(\cdot\,\bW^{(\ell)})$ defines the update function. 

This hierarchical formulation provides a unified and differentiable framework that captures multi-scale dependencies across nodes, edges, and the global graph, making GNNs particularly effective for structured learning tasks such as spatial forecasting and relational time-series prediction.

\begin{figure}[t!]
    \centering
    \scalebox{.8}{\begin{tikzpicture}[node distance=2cm]
        \begin{scope}[xshift=-1.5cm]
            \draw[fill=gray!10, rounded corners] (-0.7,-5.7) rectangle (6,1);

            \node[draw, circle, fill=gray!20] (Node_n) {$V_\ell$};
            \node[draw, circle, fill=gray!20, below of=Node_n] (Edge_n) {$E_\ell$};
            \node[draw, circle, fill=gray!20, below of=Edge_n] (Global_n) {$U_\ell$};
              
            \node[draw, circle, fill=gray!50, right=4cm of Node_n] (Node_np1) {$V_{\ell+1}$};
            \node[draw, circle, fill=gray!50, below of=Node_np1] (Edge_np1) {$E_{\ell+1}$};
            \node[draw, circle, fill=gray!50, below of=Edge_np1] (Global_np1) {$U_{\ell+1}$};
              
            \draw[->, dashed, darkblue] (Node_n) -- node[midway, above] {$\phi^v$} (Node_np1);
            \draw[->, dashed, darkblue] (Edge_n) -- node[midway, above] {$\phi^e$} (Edge_np1);
            \draw[->, dashed, darkblue] (Global_n) -- node[midway, above] {$\phi^u$} (Global_np1);

            \draw[- Circle, dotted, teal!70!black, thick] (0.5, 0) to [bend right=50] node[midway, left] {$\rho_{V_\ell\to E_\ell}$} (2, -2.);
            \draw[-, dotted] (0.5, -4) to [bend left=50](2, -2.);
            \draw[-, dotted] (0.5, -4) to [bend right=50] (4, 0.);
            \draw[- Circle, dotted, orange!70!black, thick] (2, -2) to [bend left=50] node[midway, right] {$\rho_{E_\ell\to V_\ell}$} (4, 0.);
            \draw[- Circle, dotted, green!60!black, thick] (2, -2) to [bend right=50] node[midway, right] {$\rho_{E_\ell\to U_\ell}$} (4, -4);
            \draw[-, dotted] (0.5, 0) to [bend left=50] (4, -4.);

            \node[below] at (2.5, -5) {\textit{Message passing layer}};
        \end{scope}
    
        \begin{scope}[xshift=-6cm, yshift=-3cm]
            \draw[dashed, rounded corners] (-0.7,-1) rectangle (2.7,3);
            
            \node[circle, draw, fill=blue!25, minimum size=.7cm] (Left_A) at (0,0) {};
            \node[circle, draw, fill=teal!25, minimum size=.7cm] (Left_B) at (2,0) {};
            \node[circle, draw, fill=purple!25, minimum size=.7cm] (Left_C) at (2,2) {};
            \node[circle, draw, fill=orange!25, minimum size=.7cm] (Left_D) at (0,2) {};
              
            \draw[gray!70] (Left_A) -- (Left_B);
            \draw[gray!70] (Left_B) -- (Left_C);
            \draw[gray!70] (Left_C) -- (Left_D);
            \draw[gray!70] (Left_D) -- (Left_A);
            
            \draw[-] (2.7,-1) -- (3.8,-2.6);
            \draw[-] (2.7,3) -- (3.8,3.93);
        \end{scope}
        
        \begin{scope}[xshift=6cm, yshift=-3cm]
            \draw[dashed, rounded corners] (-0.7,-1) rectangle (2.7,3);
            
            \node[circle, line width=1pt, draw, fill=blue!50, minimum size=.7cm] (Right_A) at (0,0) {};
            \node[circle, line width=1pt, draw, fill=teal!50, minimum size=.7cm] (Right_B) at (2,0) {};
            \node[circle, line width=1pt, draw, fill=purple!50, minimum size=.7cm] (Right_C) at (2,2) {};
            \node[circle, line width=1pt, draw, fill=orange!50, minimum size=.7cm] (Right_D) at (0,2) {};
              
            \draw[gray!80, line width=1.5pt] (Right_A) -- (Right_B);
            \draw[gray!80, line width=1.5pt] (Right_B) -- (Right_C);
            \draw[gray!80, line width=1.5pt] (Right_C) -- (Right_D);
            \draw[gray!80, line width=1.5pt] (Right_D) -- (Right_A);

            \draw[-] (-1.5,-2.6) -- (-0.7,-1);
            \draw[-] (-1.5,3.93) -- (-0.7,3);
        \end{scope}
    \end{tikzpicture}}
    \caption{Example of a message passing layer in a GNN. $V_\ell$, $E_\ell$ and $U_\ell$ respectively refer to node, edge, and global level at layer $\ell$. $\phi$ and $\rho$ are respectively update and propagation functions.}
    \label{fig: message_passing}
\end{figure}

\section{Methodology}\label{sec:methodo}
\subsection{Graph-based representation of electricity networks}

A central step in GNNs to electricity demand forecasting lies in the definition of an appropriate graph structure. In such representations, nodes correspond to load entities---such as regions, substations, or feeders---while edges capture spatial or statistical dependencies between them. The choice of graph topology is not trivial: while physical grids are naturally graph-structured, the objective in forecasting is not to reproduce the transmission network itself, but rather to construct a representation that best reflects the correlations driving demand patterns.
An 
intuitive strategy is to construct graphs based on spatial proximity:  
the connection strength between two nodes, $u$ and $v$ is given by the geodesic distance $\mathbf{dist}(u,v)$, kernelized through a Gaussian kernel as $\mathbf{d}(u,v) = \exp\left\{-\frac{\mathbf{dist}(u,v)^2}{\sigma_b^2}\right\}$, where $\sigma_b$ is the kernel bandwidth controlling the rate of decay. The final weighted adjacency matrix is then given by thresholding $\mathbf{d}(u,v)$ by a given level $\lambda$:
\begin{align}
    \boldsymbol{W}_\lambda = (\boldsymbol{W}_{uv})_{u,v \in \V} = \mathbf{d}(u,v) \cdot \Ind{\mathbf{d}(u,v) \geq \lambda}.
\end{align}
In practice, $\lambda$ is picked so that the graph induced by $\boldsymbol{W}_\lambda$ is minimally connected, and $\sigma_b$ is set equal to the median pairwise geodesic distance. By tuning the kernel bandwidth and connectivity threshold, one obtains sparse yet connected graphs that reflect the spatial contiguity of administrative or physical regions \citep{williams2020electricity}. Although simple, this approach may overlook structural barriers across the terrain, e.g. mountains or coasts, which can alter demand correlations.

A second family of methods derives graphs from meteorological and load signals. Regional load curves or weather time-series (e.g. temperature, cloud cover, wind speed) can be aggregated and compared across nodes using distance or similarity measures. Techniques such as Dynamic Time Warping (DTW), correlation matrices, or precision matrices (inverse covariance) provide weighted adjacency structures that emphasize regions exhibiting similar load responses to external drivers. Compared to purely geographical graphs, these data-driven constructions offer the advantage of capturing latent dependencies 
beyond physical distance.
Hybrid strategies have also been considered, combining spatial kernels with correlation-based graphs or spectral methods such as GL3SR \citep{humbert2021learning}, in order to balance physical interpretability with statistical efficiency. Synthetic datasets, where spatial and temporal correlations are artificially imposed, further allow benchmarking the ability of these constructions to recover meaningful edges before applying them to real data.
In this work, we adopt a comparative perspective, evaluating multiple graph inference strategies---geographical, correlation-based, and hybrid---on French regional and UK residential datasets. This enables us to assess not only their impact on forecasting performance but also their contribution to the interpretability of learned spatial dependencies.

\subsection{Graph Neural Networks architectures}
Once a suitable graph structure has been defined, the next step is to select appropriate Graph Neural Network (GNN) architectures for forecasting tasks. GNNs operate by iteratively propagating information across the nodes of a graph through a process known as message passing. At each layer, a node updates its representation by aggregating information from its neighbors and combining it with its own features. Stacking multiple layers extends the receptive field, allowing the model to capture interactions at increasingly larger scales. This approach 
can be 
seen as an extension of convolution to non-Euclidean domains, tailored to exploit relational inductive biases in the data.
Several GNN variants have been proposed in the literature, differing primarily in their aggregation and propagation mechanisms:

\begin{itemize}
    \item \textbf{Graph Convolutional Networks} (GCNs). 
		GCNs generalize the classical convolution operator to graphs by normalizing and aggregating neighboring features \cite{kipf2017semisupervised}. They provide a computationally efficient baseline and remain competitive in many forecasting tasks, though their fixed aggregation scheme limits expressiveness;
    \item \textbf{GraphSAGE}. The Graph SAmple and aggreGatE framework \citep{hamilton2017inductive} extends GCNs by learning inductive functions that generate node embeddings from sampled neighborhoods. The choice of aggregation function (mean, max-pooling, LSTM) controls the balance between model complexity and flexibility. GraphSAGE is particularly well suited to settings where new nodes may appear over time, a common occurrence in dynamic energy networks;
    \item \textbf{Approximate Personalized Propagation of Neural Predictions} (APPNP). APPNP \citep{gasteiger2018predict} decouples feature transformation from information diffusion by applying a personalized PageRank scheme \citep{page1999pagerank}. This design mitigates over-smoothing and provides a stable way to propagate information across multiple hops, making it robust in scenarios with sparse or noisy data;
    \item \textbf{Graph Attention Networks} (GATs). Building on attention mechanisms \citep{velivckovic2018graph}, GATs assign learnable weights to each edge during aggregation, allowing the model to focus selectively on the most relevant neighbors. This flexibility enhances performance on heterogeneous graphs and provides interpretable attention scores that can reveal evolving interaction patterns. Recent extensions such as GATv2 \citep{brody2021attentive} and TransformerConv \citep{shi2020masked} further increase expressiveness by addressing limitations of the original formulation;
    \item \textbf{Spectral and topology-adaptive models}. Architectures such as ChebConv \citep{defferrard2016convolutional} or TAGConv \citep{du2017topology} leverage polynomial filters of the graph Laplacian to capture information at multiple neighborhood depths. These methods provide a flexible trade-off between spectral and spatial approaches, particularly effective when long-range dependencies are important.
\end{itemize}
In this study, we benchmark a wide spectrum of these GNN architectures. By comparing convolution-based, attention-based, and diffusion-based models, we aim to disentangle the trade-offs between predictive accuracy, robustness, and interpretability in the context of electricity demand forecasting.

\subsection{Interpretability}
Forecasting models for energy applications must not only achieve highly accurate predictions, but also provide insights about the factors shaping their predictions. Interpretability is crucial for operational adoption, as system operators require models that can be audited, trusted, and are aligned with expert knowledge. In the context of GNNs, there are several complementary approaches for this objective.

\paragraph{GNNExplainer}
A widely used approach is GNNExplainer \citep{ying2019gnnexplainer}, which identifies the subgraphs and node features that influence a model’s decision the most. The method optimizes the mutual information between the forecast and a compact explanatory subgraph, distinguishing relevant from irrelevant connections. In practice, this produces a mask over the adjacency matrix and feature space, highlighting which regional links and covariates contribute most to the prediction. For electricity demand, GNNExplainer can uncover correlations between areas (e.g. climatic zones with similar load behaviors) that are not explicitly encoded in the input graph.

\paragraph{Attention-based interpretability}
Attention mechanisms in GATs provide another perspective by assigning learnable weights to neighbors during message passing. These coefficients quantify the relative importance of edges and evolve dynamically over time. Since raw attention scores may not always provide reliable explanations \citep{jain2019attention}, we complement them with dimensionality reduction: sequences of attention matrices are projected into lower dimensions using PCA or UMAP. This makes it possible to identify recurrent spatio-temporal patterns, such as seasonal groupings of regions or structural shifts during exceptional weather conditions. The resulting projections transform complex attention dynamics into interpretable structures aligned with known system behaviors.

\paragraph{ALE plots}
At the feature level, ALE plots \citep{apley2020visualizing} offer an effective tool to assess how variations in explanatory variables affect model outputs. Unlike partial dependence methods, ALE avoids extrapolation in sparse regions of the input space, ensuring more faithful interpretations. In energy forecasting, this technique helps in quantifying the impact of factors such as temperature or cloud cover on predicted load, complementing the relational explanations obtained from GNNExplainer and attention analysis.

Taken together, these approaches---subgraph extraction (GNNExplainer), dynamic edge weighting (attention), and feature-level effects (ALE)---form a comprehensive framework for interpreting forecasts produced by GNNs. 

\subsection{Expert aggregation}
A simple yet effective way to quantify uncertainty in neural networks is to train multiple models independently with different random initializations, and then aggregate their predictions \citep{lakshminarayanan2017simple}. This approach, known as Deep Ensembles, captures epistemic uncertainty ---i.e. uncertainty stemming from limited knowledge or data--- by leveraging the variability in predictions induced by different parameter configurations. It is particularly well-
suited when models exhibit high variance, than can happen if small amounts of data are available or the variables are highly correlated. Given that each forecasting model carries distinct strengths and inductive biases, we further exploit their complementarity through expert aggregation. In particular, we employ the ML-Poly algorithm \citep{gaillard2014second}, a robust online learning method that dynamically adjusts expert weights based on past forecasting performance. Formally, for time step $t$, let $x_{j,t}$ denote the forecast of expert $j$ and $p_{j,t}$ its weight; the aggregated prediction is given by $\widehat y_t = \sum_{j=1}^K p_{j,t} x_{j,t}$
where $K$ is the number of experts. To avoid overfitting and reduce model selection complexity, we also consider uniform aggregation that simply averages over all expert forecasts. Despite its simplicity, this baseline often performs competitively ---sometimes even outperforming the best individual expert--- thanks to the regularizing effect of ensemble averaging which reduces the variance of predictions.

\section{Experiments}
This section details the experimental design used to evaluate the proposed graph-based forecasting framework. We describe the datasets, data preprocessing and normalization, the training procedures, and the benchmark models against which the GNN architectures are compared.

\subsection{Datasets}
\paragraph{French Regional Load Dataset}
We evaluate GNN models on regional electricity load data provided by RTE (Réseau de Transport d'Électricité), the French electricity transmission system operator. The dataset spans from January $2015$ to December $2019$ and includes various features relevant to electricity demand and meteorological conditions across French administrative regions. Calendar-related features include a trend index, multiple date encodings, month and year indicators, time-of-day and day-of-year information, week numbers, day type (weekday, weekend, public holiday), specific holiday periods (e.g. Christmas, summer, national holidays), and a daylight saving time indicator. Meteorological features (e.g. temperature, wind, solar irradiance) are averaged across the $32$ national weather stations and spatially aggregated by administrative region, providing region-level summaries of environmental conditions. The target variable is the national electricity load (in megawatts). We also include lagged load features, as detailed in Sec.~\ref{sec:exp-setting}.

\paragraph{UK Power Networks Dataset}
The UK dataset\footnote{\url{https://weave.energy/}} consists of aggregated half-hourly residential electricity load data collected via smart meters by four UK Distribution Network Operators (DNOs). Currently, data are available from the two largest DNOs, Scottish and Southern Electricity Networks (SSEN) and National Grid Electricity Distribution (NGED), covering approximately $2$ million smart meters, $120,000$ low voltage feeders, and $50,000$ substations, starting from January $2024$ (NGED) and February $2024$ (SSEN), with updates subject to publication delays of $5$ to $30$ days. The data primarily cover regions such as Northern Scotland, the Midlands, South Wales, South West England, and Southern England. In our experiments, we use only calendar-based features (e.g. day of the week, time of day) and lagged load values, consistent with our setup for the French dataset. The target variable corresponds to the electricity demand measured at residential substations.

\paragraph{Synthetic Datasets}
To systematically evaluate the performance and interpretability of graph neural networks (GNNs), we generate synthetic datasets with controlled spatial and temporal dependencies. This approach allows us to establish a lower bound on model performance and to investigate attention mechanisms in a controlled setting.
In the first type of synthetic data, node-level temperature and load series are generated to mimic real-world variability. Let $T_j^\text{gen}(t)$ and $L_j^\text{gen}(t)$ denote the generated temperature and load at node $j$ and time $t$, with $T_j^\text{gen}(t) = a t + b_j (\cos \omega_1 t + \cos \omega_2 t)$, where $a$ is a linear trend coefficient, $\omega_1$ and $\omega_2$ encode daily and yearly cycles, and $\mathbf b = (b_j)_{1 \le j \le N} \sim \mathcal{N}(\hat{\boldsymbol{\mu}}, \hat{\mathbf{C}})$ with empirical mean $\hat{\boldsymbol{\mu}}$ and covariance $\hat{\mathbf{C}}$ estimated from observed temperatures. Temperatures are rescaled to match observed regional minima and maxima. Loads are then derived via a cubic spline mapping $L_j^\text{gen}(t) = \tilde f_j(T_j^\text{gen})(t) + \varepsilon_j(t)$, with $\tilde f_j$ fitted to the observed data and $\boldsymbol{\varepsilon} \sim \mathcal{N}(\mathbf 0, \mathbf \Sigma)$. Two covariance structures are considered: $\mathbf \Sigma = \mathbf{I}$ for independent regions, and $\mathbf \Sigma = \boldsymbol{\rho}(\mathbf W_\lambda)$ to model pairwise spatial interactions.
In the second type, designed to study attention mechanisms, each dataset comprises $N$ node-level time-series formed by a smooth trend, a periodic component, and strong high-frequency noise to prevent trivial per-node fitting. Node interactions are introduced by multiplying the input matrix $\mathbf X \in \mathbb{R}^{N \times T}$ with a coupling matrix $\mathbf A \in \mathbb{R}^{N \times N}$. Three variations are generated:
\begin{itemize}[topsep=0.1em]
\item \textit{Single coupling}: a fixed, sparse, full-rank matrix $\mathbf A$ is applied at all time steps.
\item \textit{Explicit switching}: two distinct matrices $\mathbf A_1$ and $\mathbf A_2$ are alternated according to a binary switching signal provided as an additional input.
\item \textit{Ambiguous switching}: the same alternation occurs, but two categorical inputs encode the active coupling only jointly, mimicking latent factors that are not directly observable.
\end{itemize}
The supervised task is to reconstruct the original signals, ensuring that predictive performance relies on leveraging the inter-node dependencies encoded in $\mathbf A$.

\subsection{Experimental settings}\label{sec:exp-setting}
We conduct our experiments within a supervised regression framework, where the objective is to forecast electricity load at multiple nodes of a power network using temporal and meteorological features. Two complementary forecasting settings are considered, corresponding to the experimental frameworks of our previous studies.

\paragraph{Single-step regression setup}  
This corresponds to a \textit{direct regression} approach, commonly adopted in electricity load forecasting when the emphasis is placed on exogenous explanatory factors rather than temporal autocorrelation. Electricity demand at each time step $t$ is forecast based solely on exogenous covariates observed at the same time. This corresponds to a pointwise regression problem, where the model learns a mapping $\Phi_{\boldsymbol{\theta}} : (\mathcal{G}, \tilde{\mathbf{X}}_t) \mapsto \hat{\mathbf{y}}_t$, with $\tilde{\mathbf{X}}_t \in \mathbb{R}^{N\times d}$ representing normalized features for all $N$ nodes and $d$ covariates. The features include calendar indicators, weather-related variables, and regional characteristics, allowing the model to capture spatial dependencies encoded in the graph structure $\mathcal{G}$. The model parameters $\boldsymbol{\theta}$ are optimized by minimizing the 
MSE between predicted and observed loads. 

\paragraph{Sequence-to-sequence forecasting setup}  
This 
extends the previous formulation to a temporal prediction task, where the model is trained to forecast the next day’s load profile based on the preceding day’s data. Specifically, each input sequence covers the past $48$ half-hourly intervals (i.e. $24$ hours), and the model predicts the load values for the following $48$ intervals. The forecasting task is thus framed as a sequence-to-sequence regression problem:
\[
\hat{\mathbf{y}}_{t:t+h-1} = \Phi_{\boldsymbol{\theta}}(\mathcal{G}, \tilde{\mathbf{X}}_t \cup \mathbf{y}_{t-w:t-1}),
\]
where $w = h = 48$. This formulation enables the model to capture both spatial dependencies (via the graph structure) and temporal dependencies (through the inclusion of lagged load values). It also better reflects real-world operational settings where short-term load forecasts are required based on recent historical patterns.

\paragraph{Training Procedure}  
In both setups, the feature tensor $\tilde{\mathbf{X}} \in \mathbb{R}^{N \times d \times T}$ and the corresponding targets $\mathbf{y} \in \mathbb{R}^{N \times T}$ are normalized to the range $[0,1]$, both nodewise and channelwise. Parameters are optimized by minimizing the MSE loss using the Adam optimizer \citep{kingma2014adam}, with early stopping based on validation error. For the French dataset, the training, validation, and test periods respectively span $2015$--$2017$, $2018$, and $2019$. For the UK dataset, we adopt a shorter temporal window: training from February $13$--$21$, $2024$, validation on February $22$, and testing from February $23$--$25$, $2024$.  
The models are trained using mini-batch stochastic gradient descent with non-overlapping input/output windows to ensure independence between training samples.











\begin{figure}[t!]
    \centering
    \scalebox{1}{
    \begin{tikzpicture}[font=\small, node distance=0.5cm and 0.5cm]
        \node at (4.5, 2.3) {\textbf{French Dataset}};
        \draw[->, thick] (-2,1.5) -- (12,1.5);

        \fill[blue!20] (-2,1.3) rectangle (5.56,1.7);
        \node at (1.78,1.5) {Train};

        \fill[orange!20] (5.56,1.3) rectangle (8.08,1.7);
        \node at (6.82,1.5) {Val};

        \fill[green!20] (8.08,1.3) rectangle (10.6,1.7);
        \node at (9.34,1.5) {Test};

        \draw (-2,1.3) -- (-2,1.7);
        \draw (5.56,1.3) -- (5.56,1.7);
        \draw (8.08,1.3) -- (8.08,1.7);
        \draw (10.6,1.3) -- (10.6,1.7);

        \node at (-2,1.1) {\footnotesize Jan '15};
        \node at (5.56,1.1) {\footnotesize Jan '18};
        \node at (8.08,1.1) {\footnotesize Jan '19};
        \node at (10.6,1.1) {\footnotesize Jan '20};

        \node at (4.5, .3) {\textbf{UK Dataset}};
        \draw[->, thick] (-2, -0.5) -- (12, -0.5);

        \fill[blue!20] (-2,-0.7) rectangle (7,-0.3);
        \node at (2.5,-0.5) {Train};

        \fill[orange!20] (7.,-.7) rectangle (7.9,-0.3);
        \node at (7.45,-0.5) {Val};

        \fill[green!20] (7.9,-.7) rectangle (10.6,-0.3);
        \node at (9.15,-0.5) {Test};

        \draw (-2,-.7) -- (-2,-0.3);
        \draw (7.,-.7) -- (7.,-0.3);
        \draw (7.9,-.7) -- (7.9,-0.3);
        \draw (10.6,-.7) -- (10.6,-0.3);

        \node at (-2,-0.9) {\footnotesize Feb 13};
        \node at (6.9,-0.9) {\footnotesize Feb 22};
        \node at (8.1,-0.9) {\footnotesize Feb 23};
        \node at (10.6,-0.9) {\footnotesize Feb 26};
    \end{tikzpicture}}

    \caption{Temporal splits for training, validation, and testing on the French and UK datasets.}
    \label{fig:dataset-splits}
\end{figure}

\paragraph{Evaluation Metrics}  
Forecasting accuracy is evaluated using two complementary metrics: the Mean Absolute Percentage Error (MAPE) and the Root Mean Square Error (RMSE).
\begin{equation}
\text{MAPE}(\mathbf{y}, \hat{\mathbf{y}}) = \frac{1}{T} \sum_{t=1}^T \left| \frac{\sum_{i=1}^N y_{i,t} - \hat{y}_{i,t}}{\sum_{i=1}^N y_{i,t}} \right|, 
\qquad 
\text{RMSE}(\mathbf{y}, \hat{\mathbf{y}}) = \sqrt{\frac{1}{T}\sum_{t=1}^T \left(\sum_{i=1}^N y_{i,t} - \hat{y}_{i,t}\right)^2}.
\end{equation}
MAPE offers an interpretable percentage-based measure of relative error, facilitating comparison across datasets and scales, while RMSE penalizes large deviations and is thus more sensitive to peak load mispredictions \citep{hyndman2006another}.  
Together, the two metrics provide a balanced evaluation of 
accuracy and robustness, consistent with best practices in the 
forecasting literature.

\paragraph{Baseline Models}
In the context of simple regression, we compared basic GNNs (GCN \& SAGE) with GAMs. In the sequence-to-sequence forecasting case, we evaluated three baseline forecasting strategies to predict the aggregated electricity demand at multiple load nodes. 

The first baseline approach trains a separate feedforward neural network for each node to forecast the next $48$ half-hourly values (i.e. the next day), using all available information up to time $t - 1$. 
The second baseline is a simple yet competitive persistence method, where the forecast for each time slot is directly copied from the same half-hour either on the previous day ($D-1$) or the previous week ($D-7$), a common benchmark in load forecasting studies \citep{hyndman2010density}.
The third baseline utilizes TiREX \citep{auer2025tirex}, a state-of-the-art foundation model pre-trained on a vast corpus of time-series, which has demonstrated top-ranking performance on energy benchmarks such as \texttt{gifteval} at the time of this study. We implement two variations of this model, both without using exogenous covariates: (i) a bottom-up approach that independently forecasts the next $48$ half-hourly values for each node, and (ii) an aggregated approach that forecasts the same horizon for the sum of all nodes. These baselines provide a diverse set of references spanning machine learning, statistical, and naive forecasting paradigms. While GAMs are widely used in electricity forecasting for their ability to model nonlinear effects of multiple covariates, we did not include them here due to the limited availability of informative exogenous features in our setting ---particularly in the UK dataset, where only calendar features and lagged load values are available. In such contexts, the strength of GAMs in capturing smooth covariate effects becomes less relevant, and simpler autoregressive strategies often suffice \citep{haben2018short}.

\paragraph{Parametrization}
To ensure fair and consistent comparisons across the various graph-based architectures, model hyperparameters were optimized using the Optuna library \citep{akiba2019optuna}. The search space included core architectural and training parameters common to all models: the number of layers (\texttt{n\_layers}), hidden channels per layer (\texttt{hidden\_channels}), batch size (\texttt{batch\_size}), and learning rate (\texttt{lr}). Each parameter plays a distinct functional role --- \texttt{n\_layers} determines the model depth and receptive field, \texttt{hidden\_channels} regulates representational capacity, \texttt{batch\_size} impacts training stability and generalization, and \texttt{lr} governs convergence dynamics.  Model-specific hyperparameters were also tuned to account for architectural diversity. For spectral and spatial models such as ChebConv and TAGConv, we optimized the polynomial order or neighborhood size, which control the locality of information propagation. For attention-based architectures (GAT, GATv2, TransformerConv), we tuned the number of attention heads to modulate the diversity of subspace projections. Finally, for diffusion-based models such as APPNP, we optimized the number of propagation steps $K$ and the teleportation probability $\alpha$, which jointly balance local and global signal diffusion.  All experiments employed early stopping based on the validation loss, with each configuration evaluated through multiple random seeds to ensure robustness. The selected hyperparameters are reported in Table~\ref{table: gs1and2}, and correspond to the configurations yielding the lowest validation RMSE. Notably, graph construction strategies were adapted to each dataset: in the French case, regional aggregation made spatially informed graphs (based on proximity or meteorological similarity) meaningful, whereas for the UK data, data-driven graphs proved more suitable since that dataset exhibits finer spatial granularity and noisier correlations.

\setlength{\aboverulesep}{0pt}
\setlength{\belowrulesep}{1.5pt}

\begin{table}[t!]
    \caption{Hyperparameters of the GNN models for the French and the UK dataset in the sequence-to-sequence forecasting setup.}
    \label{table: gs1and2}
    \centering\footnotesize
    \scalebox{0.8}{
    \begin{tabular}{cccccccc}
        \toprule
        Dataset & Model & Graph structure & \texttt{batch\_size} & \texttt{n\_layers} & \texttt{hidden\_channels} & \texttt{lr} & Other HP \\        
        \midrule
        \multirow{8}{*}{\worldflag[length=2cm, width=1.4cm, stretch=0, framewidth=0mm]{FR}}
            & GCN & Space & 16 & 1 & 170 & $3\cdot10^{-3}$ & None \\
            & SAGE & Correlation & 16 & 2 & 401 & $4\cdot10^{-4}$ & None \\
            & GAT & GL3SR & 16 & 3 & 364 & $1\cdot10^{-3}$ & $\texttt{heads}=2$ \\
            & GATv2 & Space & 16 & 2 & 359 & $4\cdot10^{-4}$ & $\texttt{heads}=2$ \\
            & Transformer & DistSplines & 16 & 1 & 335 & $3\cdot10^{-3}$ & $\texttt{heads}=2$ \\
            & TAG & DTW & 16 & 3 & 329 & $3\cdot10^{-4}$ & $K=1$ \\
            & Cheby & DistSplines & 16 & 1 & 249 & $6\cdot10^{-4}$ & $K=3$ \\
            & APPNP & Space & 16 & 1 & 106 & $2\cdot10^{-3}$ & $(K, \alpha)=(1,0.93)$ \\
        \midrule\midrule
        \multirow{8}{*}{\worldflag[length=2cm, width=1.4cm, stretch=0, framewidth=0mm]{GB}}
            & GCN & Correlation & 128 & 1 & 96 & $5\cdot10^{-3}$ & None \\
            & SAGE & Correlation & 32 & 5 & 207 & $7\cdot10^{-4}$ & None \\
            & GAT & Correlation & 16 & 1 & 172 & $3\cdot10^{-3}$ & $\texttt{heads}=1$ \\
            & GATv2 & Precision & 128 & 3 & 71 & $2\cdot10^{-3}$ & $\texttt{heads}=1$ \\
            & Transformer & Correlation & 32 & 5 & 354 & $4\cdot10^{-4}$ & $\texttt{heads}=1$ \\
            & TAG & DTW & 64 & 1 & 70 & $3\cdot10^{-3}$ & $K=4$ \\
            & Cheby & Correlation & 32 & 4 & 118 & $6\cdot10^{-2}$ & $K=10$ \\
            & APPNP & Precision & 64 & 3 & 454 & $2\cdot10^{-2}$ & $(K, \alpha)=(8,0.85)$ \\
        \bottomrule
    \end{tabular}}
\end{table}

\section{Results}

\subsection{Numerical Results}
We report here the main empirical findings obtained across 
two experimental frameworks: the single-step regression setup and the sequence-to-sequence forecasting setup. Together, these experiments shed light on how graph construction, architectural choices, and aggregation strategies influence the forecasting accuracy of GNNs under different spatial and temporal conditions.  

\paragraph{Single-step regression setup}
The single-step regression experiments primarily aim to assess the contribution of graph-based models in structured versus unstructured settings, rather than to compete on 
mere forecasting accuracy. Using both synthetic and real datasets, we investigate how the presence or absence of inter-regional dependencies influences the relative importance of GNNs within a mixture-of-experts framework. On the French national dataset, where load and meteorological variables are aggregated at the regional level, graph-based models demonstrate clear relevance. While overall performance differences between architectures remain moderate, models exploiting spatial or correlation-based connectivity systematically outperform those relying on 
pure autoregression (i.e. identity adjacency matrices). This confirms that regional interactions encode meaningful predictive structure, and that incorporating them yields measurable gains in model generalization. Synthetic experiments further isolate the effect of these dependencies. When spatial correlations are artificially introduced between regions, GNNs achieve larger performance improvements over classical non-graph baselines such as GAMs. Conversely, in uncorrelated settings, their advantage diminishes and variance increases, reflecting the absence of useful relational signals to exploit. This contrast illustrates that GNNs derive their strength not from overparameterization, but from their ability to model relevant structured relationships present in the data. 

\begin{figure}[t!]
    \centering
    \includegraphics[width=\textwidth]{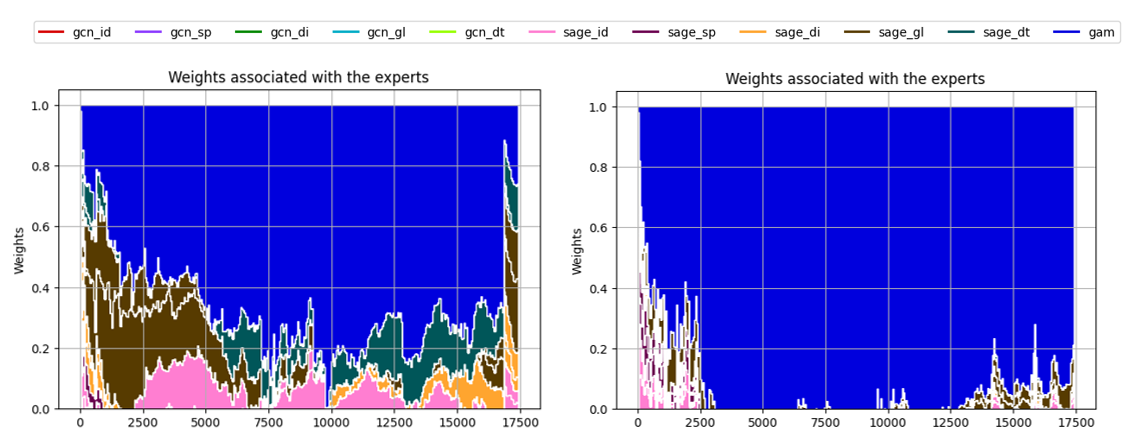}
    \caption{Weights associated with the experts on the synthetic datasets in the single-step regression setup. $\mathbf{\Sigma} = \boldsymbol{\rho}(\mathbf W_\lambda)$ (left), $\mathbf{\Sigma} = \boldsymbol{I}$ (right). GAM is the main expert, followed by mutliple SAGE-\texttt{gl3sr} and SAGE-\texttt{dtw}.}
    \label{fig: aggreg_res}
\end{figure}

The expert aggregation analysis displayed in Figure~\ref{fig: aggreg_res} reinforces this interpretation: even when their standalone performance is not dominant, GNNs contribute complementary information within the ensemble, improving the overall RMSE by approximately 200~MW compared to mixtures excluding them. Their weights within the aggregation also exhibit temporal variability, decreasing during summer and increasing in winter, which suggests that the relevance of spatial dependencies fluctuates with seasonal demand patterns. The best-performing graph constructions, obtained using the DTW similarity measure and the GL3SR graph inference approach. 

Taken together, these results emphasize that the primary value of GNNs in this setting lies in their structural interpretability and their contribution to ensemble diversity, rather than in standalone predictive dominance. When data exhibit explicit inter-node dependencies, graph-based learning becomes a principled way to integrate such information within hybrid forecasting systems.

\paragraph{Sequence-to-sequence forecasting setup}  
\begin{table}[t!]
    \centering
    \caption{Average and best test performance per model -- `Unif.' indicates uniform averaging of expert predictions, `Agg.' refers to online aggregation using the MLPol algorithm. `Bottom' aggregation aggregates at the node level before summation, while `Top' aggregation first sums expert predictions and then aggregates. `*' means the models were only ran once as as their training procedure is
    deterministic. \textbf{Bold} and \underline{underline} respectively highlight best and second per section.}
    \label{table:combined-results}
    \footnotesize
    \scalebox{.79}{
    \begin{tabular}{ccccc|cccc}
            \toprule
            \multirow{3}{*}{Model} 
            & \multicolumn{4}{c|}{French Dataset} 
            & \multicolumn{4}{c}{UK Dataset} \\
            & \multicolumn{2}{c}{MAPE (\%)} & \multicolumn{2}{c|}{RMSE (MW)} & \multicolumn{2}{c}{MAPE (\%)} & \multicolumn{2}{c}{RMSE (MW)} \\
            & Avg. $\pm$ Std & Best & Avg. $\pm$ Std & Best & Avg. $\pm$ Std & Best & Avg. $\pm$ Std & Best \\
            \midrule
            \multicolumn{9}{c}{\textbf{Baseline models}}\\
            \midrule
            FF & $1.63 \pm 0.82$ & $1.41$ & $1192 \pm 676$ & $1044$ & $14.62 \pm 0.65$ & $13.01$ & $27.27 \pm 865$ & $25.16$ \\
            \rowcolor{gray!15}TiREX (top)& $3.61^*$ & $3.61$ & $2648^*$ & $2648$ & $8.94^*$ & $8.94$ & $16.21^*$ & $16.21$\\
            TiREX (bottom)& $3.57^*$ & $3.57$ & $2633^*$ & $2633$ & $10.78^*$ & $10.78$ & $18.05^*$ & $18.05$\\
            \rowcolor{gray!15}Persistence--$1$ & $5.79^*$ & $5.79$ & $4512^*$ & $4512$ & $8.58^*$ & $8.58$ & $17.20^*$ & $17.20$ \\
            Persistence--$7$ & $6.03^*$ & $6.03$ & $4656^*$ & $4656$ & $10.03^*$ & $10.03$ & $17.21^*$ & $17.21$ \\
            \midrule
            \multicolumn{9}{c}{\textbf{Graph-based models}}\\
            \midrule
            \rowcolor{gray!15}GCN & $1.75 \pm 0.51$ & $1.21$ & $1255 \pm 359$ & $903$ & $9.52 \pm 1.39$ & $6.91$ & $ 16.68 \pm 2.27$ & $12.13$ \\
            \rowcolor{gray!15}Unif. & $1.21$ &  & $895$ &  & $8.63$ &  & $15.09$ &  \\
            \rowcolor{gray!15}Agg. & $\underline{1.09}$ &  & $\mathbf{839}$ &  & $8.51$ &  & $14.91$ &  \\
            SAGE & $1.30 \pm 0.09$ & $1.14$ & $998 \pm 72$ & $862$ & $8.63 \pm 0.84$ & $6.95$ & $15.61 \pm 1.49$ & $13.10$ \\
            Unif. & $\underline{1.12}$ &  & $\mathbf{873}$ &  & $8.19$ &  & $14.87$ &  \\
            Agg. & $1.14$ &  & $898$ &  & $\underline{8.07}$ &  & $14.85$ &  \\
            \rowcolor{gray!15}GAT & $1.42 \pm 0.23$ & $1.13$ & $1052 \pm 136$ & $888$ & $10.01 \pm 1.80$ & $6.59$ & $18.11 \pm 2.96$ & $12.79$ \\
            \rowcolor{gray!15}Unif. & $\mathbf{1.11}$ &  & $883$ &  & $\underline{7.79}$ &  & $\underline{14.13}$ &  \\
            \rowcolor{gray!15}Agg. & $\mathbf{1.08}$ &  & $\underline{871}$ &  & $8.29$ &  & $\underline{14.72}$ &  \\
            GATv2 & $1.64 \pm 0.34$ & $1.22$ & $1165 \pm 200$ & $937$ & $11.32 \pm 1.48$ & $8.21$ & $20.22 \pm 2.53$ & $15.23$ \\
            Unif. & $1.14$ &  & $888$ &  & $8.78$ &  & $16.44$ &  \\
            Agg. & $1.14$ & & $902$ &  & $8.94$ &  & $16.40$ &  \\
            \rowcolor{gray!15}Transformer & $1.50 \pm 0.17$ & $1.28$ & $1123 \pm 118$ & $950$ & $11.93 \pm 1.41$ & $9.69$ & $21.24 \pm 2.41$ & $16.61$ \\
            \rowcolor{gray!15}Unif. & $1.14$ &  & $883$ &  & $8.36$ &  & $15.32$ &  \\
            \rowcolor{gray!15}Agg. & $1.15$ &  & $910$ &  & $8.48$ &  & $15.53$ &  \\
            TAG & $1.36 \pm 0.17$ & $1.22$ & $1022 \pm 107$ & $925$ & $10.63 \pm 1.16$ & $7.69$ & $18.35 \pm 1.81 $ & $14.21$ \\
            Unif. & $1.14$ &  & $884$ &  & $9.23$ &  & $15.79$ &  \\
            Agg. & $1.14$ &  & $881$ &  & $9.41$ &  & $16.19$ &  \\
            \rowcolor{gray!15}Cheby & $1.39 \pm 0.14$ & $1.23$ & $1036 \pm 91$ & $925$ & $9.49 \pm 1.12$ & $7.95$ & $18.38 \pm 1.83$ & $15.66$ \\
            \rowcolor{gray!15}Unif. & $1.16$ &  & $\underline{878}$ &  & $8.60$ &  & $17.23$ &  \\
            \rowcolor{gray!15}Agg. & $1.15$&  & $876$ &  & $8.93$ & & $17.37$ &  \\
            APPNP & $1.48 \pm 0.17$ & $1.23$ & $1091 \pm 114$ & $931$ & $8.31 \pm 1.53$ & $6.01$ & $15.41 \pm 2.73$ & $11.01$ \\
            Unif. & $1.22$ &  & $944$ &  & $\mathbf{6.54}$ &  & $\mathbf{12.61}$ &  \\
            Agg. & $1.20$ &  & $943$ &  & $\mathbf{7.60}$ &  & $\mathbf{14.13}$ &  \\
            \midrule
            \multicolumn{9}{c}{\textbf{Aggregation of graph-based models}}\\
            \midrule
            Uniform & $1.06$ &  & $836$ &  & $8.10$ &  & $14.74$ &  \\
            Bottom Aggregation & $\mathbf{1.01}$ &  & $\mathbf{789}$ &  & $\mathbf{7.68}$ &  & $\mathbf{14.35}$ &  \\
            Top Aggregation & $1.02$ &  & $817$ &  & $8.33$ &  & $15.04$ &  \\            
            \bottomrule
        \end{tabular}}
\end{table}
When extending to a temporal forecasting horizon of $48$ half-hour intervals, we observe consistent trends across the French and UK datasets. On the French dataset, characterized by aggregated regional signals, the performance among GNN architectures is relatively homogeneous. Averaged over ten random seeds, SAGE achieves the best RMSE ($873$ MW) and a competitive MAPE ($1.12\%$), while the vanilla GAT attains the lowest MAPE ($1.11\%$), confirming its strong balance between expressiveness and stability. Simpler models such as GCN and TAG remain highly competitive, whereas more complex architectures like GATv2 and TransformerConv offer marginal or no improvements. These results suggest that, in medium-scale settings with stable aggregated targets, architectural simplicity and strong inductive biases can match or even surpass the performance of more sophisticated designs. In the more granular and noisier UK dataset, which contains residential-level signals, learning becomes more sensitive to data scarcity. In this context, APPNP achieves the best trade-off between accuracy and robustness, with the lowest MAPE ($6.54\%$) and RMSE ($12.61$ MW), followed by GAT ($7.79\%$, $14.13$ MW) and SAGE ($8.19\%$, $14.87$ MW). Complex attention-based variants such as GATv2 and TransformerConv tend to overfit, reflecting their higher parameterization relative to the available data. The strong performance of APPNP can be attributed to its diffusion mechanism with teleportation, which regularizes information propagation and mitigates overfitting. Finally, the choice of aggregation strategy plays a decisive role. On the French dataset, bottom aggregation consistently yields the best results (MAPE $=1.01\%$, RMSE $=789$ MW), suggesting that combining node-level forecasts before aggregation better captures regional patterns. On the UK dataset, where signals are heterogeneous and sparse, bottom aggregation again outperforms other strategies (MAPE $=7.68\%$, RMSE $=14.35$ MW), confirming its advantage in handling fine-grained spatial variability. In contrast, top aggregation performed after summation systematically underperforms, particularly in the UK setting. While uniform aggregation remains competitive in simpler models, adaptive 
aggregation shows systematic gains, reinforcing the benefit of data-driven combination strategies in ensemble GNN forecasting.

\smallskip
Overall, these results demonstrate that GNN-based architectures consistently outperform traditional baselines such as feedforward networks, or persistence models, and even foundation models such as TiREX, particularly in terms of RMSE. They further highlight the trade-off between model complexity and data availability: simpler GNNs (e.g. GCN, SAGE, APPNP) often achieve the best generalization, while attention-based models offer valuable interpretability without significant performance loss. This confirms the importance of balancing expressiveness, inductive bias, and data efficiency in the design of graph-based forecasting models.

\subsection{Interpretability Results}
Understanding the internal mechanisms of GNNs is essential to assess their reliability and to connect learned representations with meaningful physical or structural properties of the system under study. We therefore investigate model interpretability under both experimental setups: in the single-step regression framework, we focus on localized and feature-level explanations, while in the sequence-to-sequence setting, we analyze the attention mechanisms emerging from temporal and spatial dependencies.

\paragraph{Single-step regression setup}

\begin{figure}
    \centering
    \includegraphics[width=\linewidth]{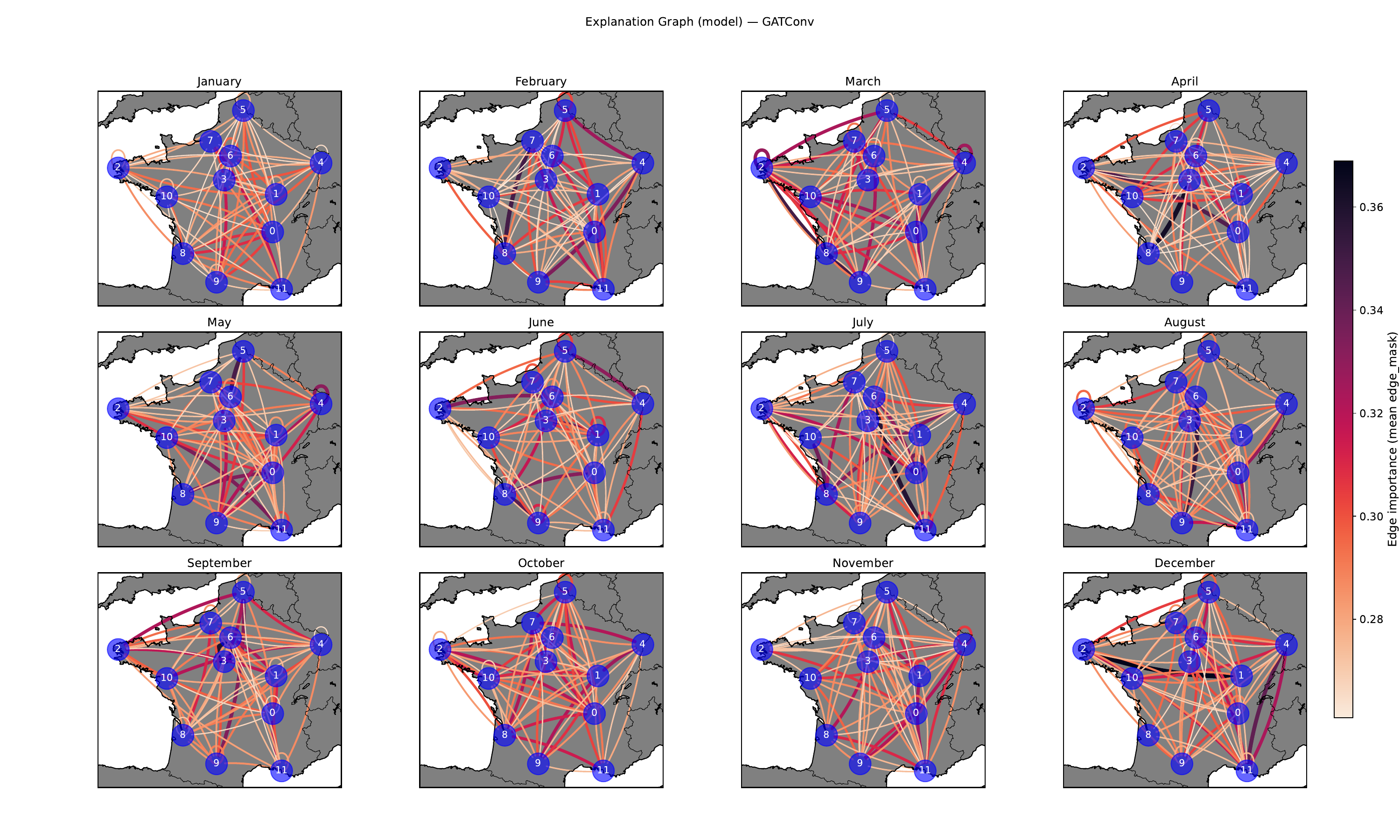}
    \caption{Explanation graphs for the GAT convolutional layer on the French dataset.}
    \label{fig:exp-graphs}
\end{figure}

\begin{figure}[t!]
    \centering
    \includegraphics[width=\textwidth]{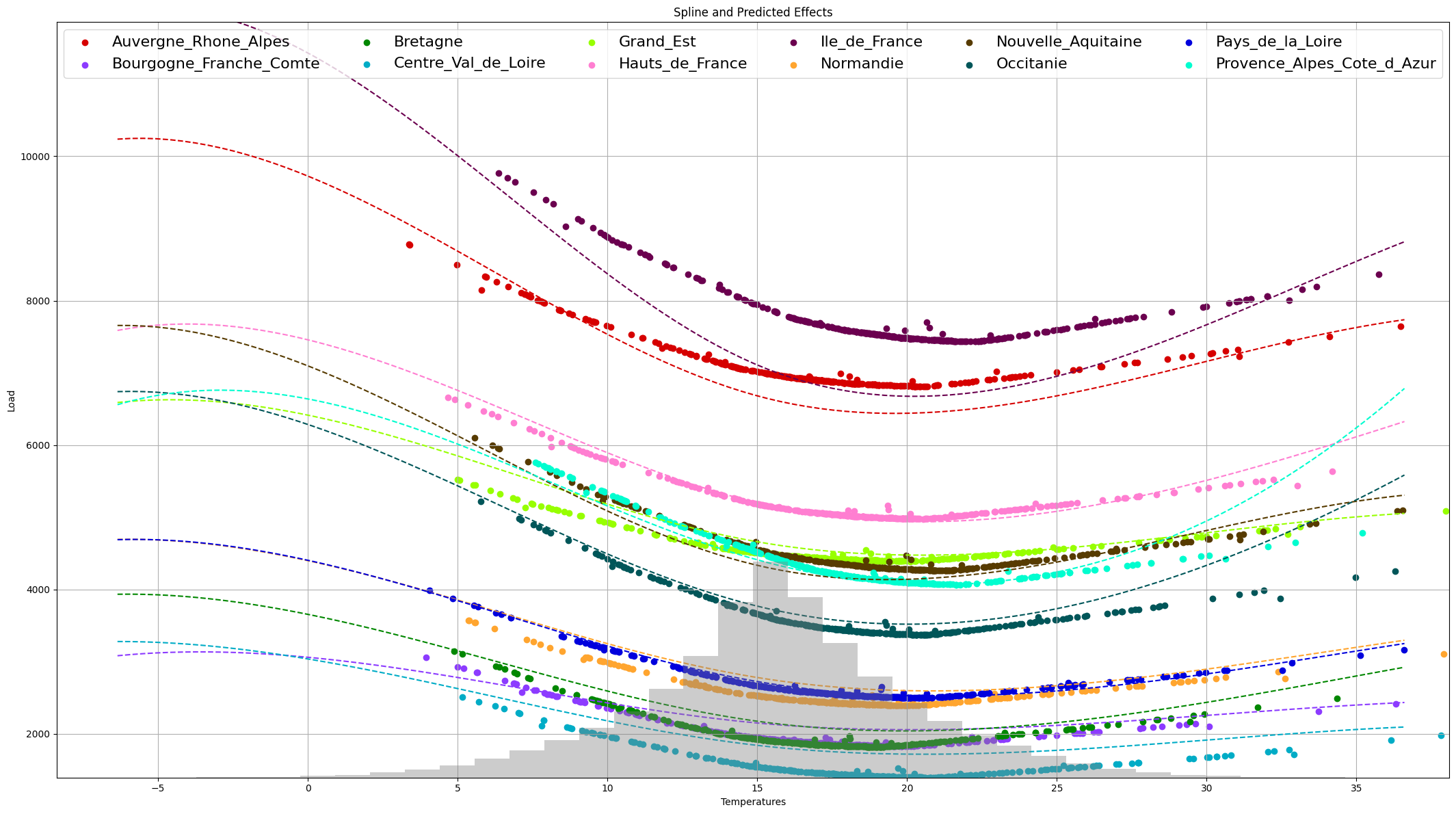}
    \caption{Spline (dashed line) and predicted (scatter plot) effects. The distribution of the generated temperatures is represented in gray.}
    \label{fig: aleplot}
\end{figure}
To interpret the behavior of graph-based models trained on the French dataset, we employ GNNExplainer \citep{ying2019gnnexplainer}, which identifies the most influential subgraphs contributing to the prediction of each region. For each season (January, June, August, November), we aggregate daily explanatory subgraphs by summing and normalizing them over all half-hour intervals. Two datasets are considered: the real regional dataset and the synthetic dataset with correlated nodes ($\boldsymbol{\Sigma} = \boldsymbol{\rho}(\mathbf{W}_\lambda)$). The learned mask $\mathbf{M}$ defines a relevance-weighted subgraph $\mathbf{W}_\lambda \odot \sigma(\mathbf{M})$, where $\odot$ denotes the Hadamard product and $\sigma$ a sigmoid normalization mapping values to $[0,1]^{N \times N}$. The resulting subgraphs reveal strong and interpretable spatial patterns. Connections between coastal regions along the Atlantic seaboard, as well as between geographically close regions such as Hauts-de-France, Normandy, and Île-de-France, emerge consistently across periods. These results indicate that the autoregressive (self-dependent) component is less dominant than initially expected, and that neighboring representations play a decisive role in explaining load variations. Interestingly, regions with lower load levels, such as Centre-Val-de-Loire, show higher variability in their explanatory weights, reflecting greater prediction uncertainty.  

For the synthetic datasets, we complement graph-level interpretability with feature-level analysis through Accumulated Local Effects (ALE) plots \citep{apley2020visualizing}. These reveal smooth and monotonic relations between temperature and load, as expected from the data generation process, but also highlight nonlinear thresholds that vary across nodes, consistent with the imposed correlation structure. Together, GNNExplainer and ALE analyses demonstrate that GNNs can recover both spatial and functional dependencies consistent with domain knowledge.

\paragraph{Sequence-to-sequence forecasting setup}  

\begin{figure}[ht]
    \centering
    \scalebox{.85}{
    \begin{subfigure}[b]{0.5\linewidth}
        \centering
        \includegraphics[width=\linewidth]{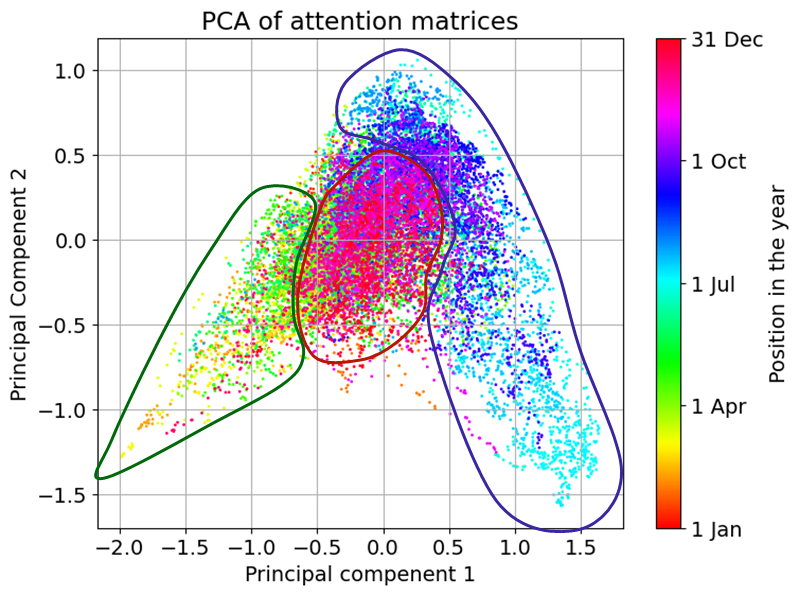}
        \caption{PCA.}
        \label{fig:pca_real_data}
    \end{subfigure}
    \hfill
    \begin{subfigure}[b]{0.38\linewidth}
        \centering
        \includegraphics[width=\linewidth]{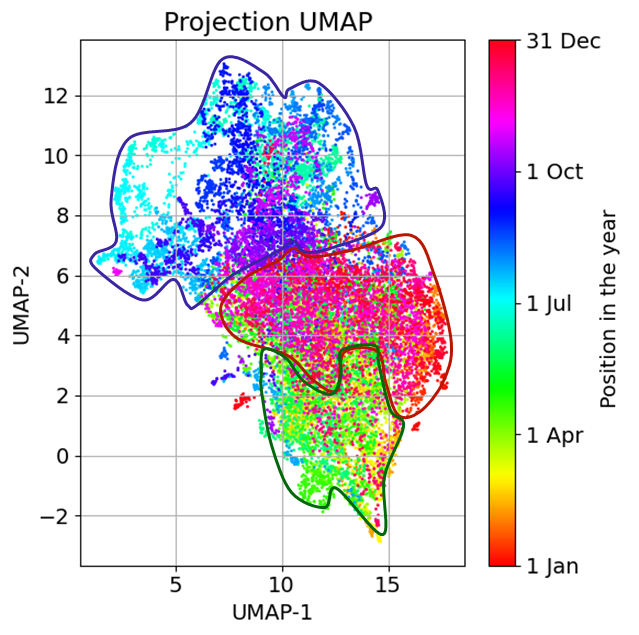}
        \caption{UMAP.}
        \label{fig:umap_real_data}
    \end{subfigure}}
		\vspace{-0.5em}
    \caption{PCA and UMAP projections of attention matrices on the French dataset reveal seasonal clustering: spring (green), summer (blue), and winter (red).}
    \label{fig:pca_umap_comparison}
\end{figure}

In the multi-step setting, we leverage the internal attention mechanisms of GAT-based architectures to gain insight into how temporal and spatial dependencies are encoded. For each time step, the model computes a full attention matrix describing the relative importance of inter-regional interactions for forecasting the next day’s load. To analyze the latent structure of these matrices, we apply dimensionality reduction techniques---Principal Component Analysis (PCA) and Uniform Manifold Approximation and Projection (UMAP)---which respectively capture linear variance and non-linear manifold structure. As shown in Figure~\ref{fig:pca_umap_comparison}, both projections reveal a clear seasonal organization: clusters corresponding to winter, summer, and transitional periods emerge naturally, indicating that the attention mechanism implicitly tracks physical drivers such as temperature, daylight variation, and regional activity cycles. 

\begin{figure}[ht!]
	\centering    
		\begin{subfigure}[b]{1\textwidth}
			\centering 
			\includegraphics[width=0.32\linewidth]{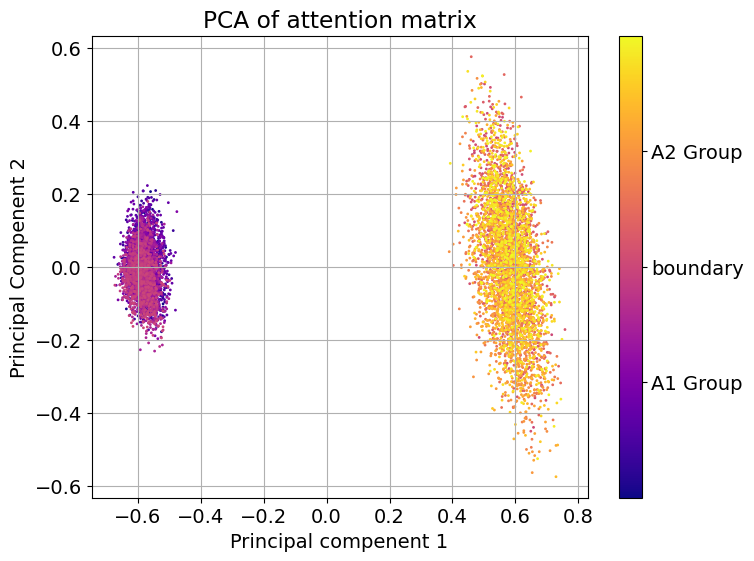}
			\caption{Explicit coupling: Layer $2$.}
			\label{fig:explicit-coupling-pca}
    \end{subfigure}\\
		\vspace{1em}
    \centering
    \begin{subfigure}[b]{0.32\textwidth}
        \centering
        \includegraphics[width=\linewidth]{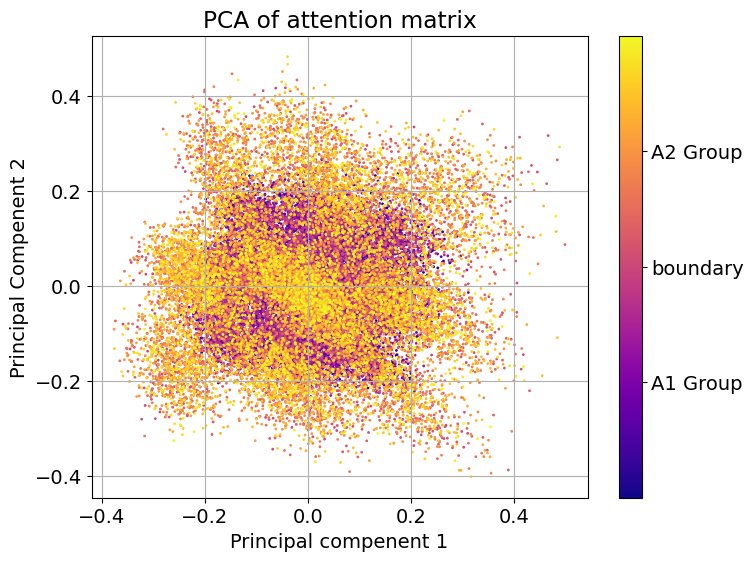}
        \caption{Implicit coupling: Layer $0$.}
    \end{subfigure}
    \hfill
    \begin{subfigure}[b]{0.32\textwidth}
        \centering
        \includegraphics[width=\linewidth]{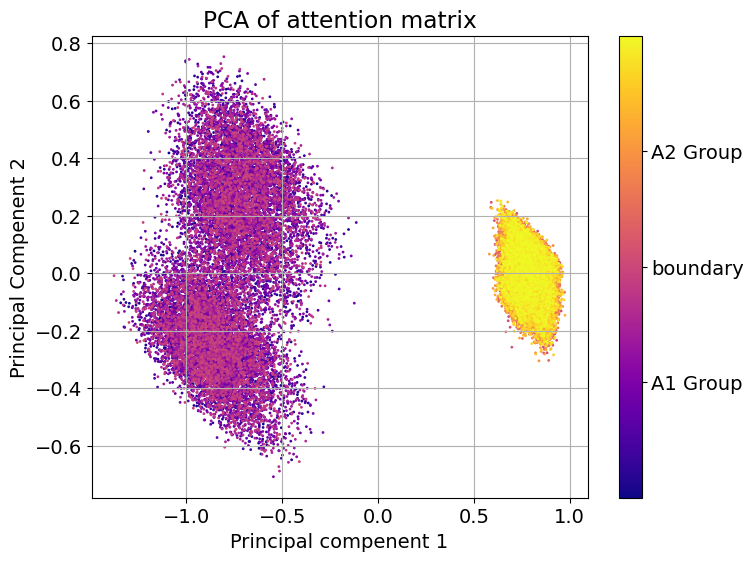}
        \caption{Implicit coupling: Layer $1$.}
    \end{subfigure}
    \hfill
    \begin{subfigure}[b]{0.32\textwidth}
        \centering
        \includegraphics[width=\linewidth]{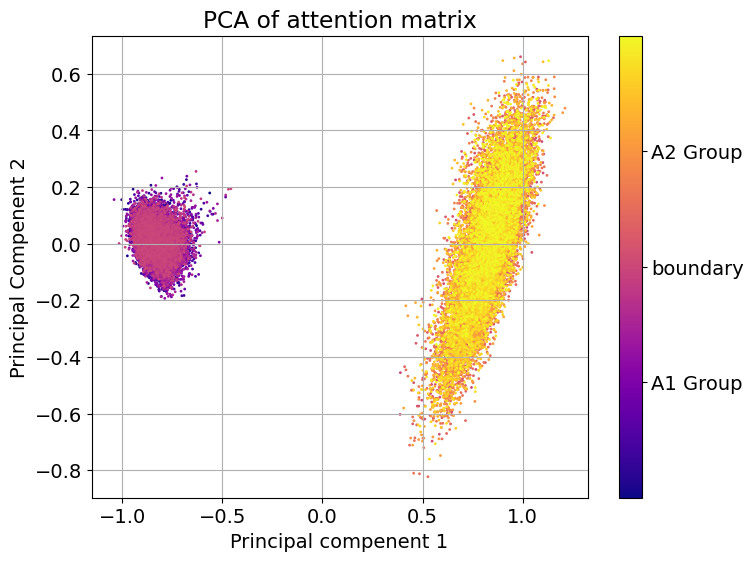}
        \caption{Implicit coupling: Layer $2$.}
    \end{subfigure}
    \caption{PCA of attention matrices on the synthetic dataset with two coupling modes ($\bA_1$ or $\bA_2$). Colors reflect the coupling matrix in effect. (a)~Explicit coupling information: the mode is explicitly given to the model as a binary input. (b---d)~Implicit coupling information: the mode in effect is indicated implicitly by the combined values of two input binary variables. The subfigures show the PCA of attention matrices across GAT layers.}
    \label{fig:PCA_hidden_coupling}
\end{figure}

We further evaluate interpretability on synthetic datasets with known coupling matrices. In the single-coupling setting, attention weights remain stable over time but do not directly mirror the true adjacency, confirming that attention scores capture task-relevant rather than purely physical dependencies. In the explicit switching scenario, where two coupling matrices alternate and the active mode is indicated by a binary variable, PCA projections of attention vectors form two well-separated clusters, showing that GATs can adapt their internal representations to changing interaction regimes. Finally, in the ambiguous switching scenario---where the active coupling is only indirectly encoded by categorical inputs---attention distributions progressively disentangle the latent regimes across layers: the first layer shows mixed clusters, intermediate layers partially separate them, and the final layer recovers the two hidden modes. This hierarchical refinement demonstrates the ability of attention-based GNNs to infer underlying structure even when exogenous information is noisy or incomplete. In real-world energy forecasting, such capacity is crucial, as many contextual variables (e.g. weather indicators, behavioral patterns) are imperfect proxies of the true generative factors. Overall, these findings highlight the complementarity of local and structural interpretability tools: while GNNExplainer and ALE reveal node-level and feature-level effects, attention analysis uncovers global, dynamic organization in learned representations, providing valuable insights into both model behavior and the underlying energy system.

\section{Conclusion}

This work investigated the use of Graph Neural Networks for electricity load forecasting, combining methodological developments and interpretability analysis. We proposed strategies for constructing and refining graph structures tailored to regional energy data, showing that incorporating spatial and correlation-based dependencies significantly enhances forecasting accuracy compared to purely autoregressive approaches. Beyond predictive performance, our results underline the interpretative value of graph-based models. On synthetic and real datasets, explanatory tools such as GNNExplainer, Accumulated Local Effects, and attention-weight analyses reveal that GNNs can recover meaningful spatial and temporal patterns, consistent with domain knowledge about regional and seasonal variability. In particular, attention mechanisms provide a structured lens to analyze evolving dependencies between regions, thereby bridging predictive modeling and physical interpretability. From a modeling perspective, our findings highlight the importance of architectural simplicity and strong inductive biases. Models such as SAGE or APPNP achieve competitive or superior performance relative to more complex architectures like GATv2 or TransformerConv, particularly in data-scarce and heterogeneous contexts. Moreover, ensemble techniques (notably bottom-up aggregations) consistently improve robustness and generalization across datasets. Despite this work remained focused on the significant problem of electricity load forecasting, which required also certain choices in the experimental setup, we believe a great part of our benchmarking framework and empirical findings can be also relevant to other spatio-temporal forecasting problems.

\paragraph{Limitations and perspectives}
Despite these encouraging results, a 
limitation of the present framework lies in the use of explicitly static graph structures. In real-world power systems, spatial and statistical dependencies evolve continuously due to changing load behaviors, demand-side flexibility, and meteorological variability. Although the attention mechanisms employed in our models already introduce a form of implicit adaptivity (by dynamically reweighting the influence of neighboring nodes over time) the underlying adjacency matrices remain fixed throughout training and inference. This static backbone may still underrepresent structural regime changes, for instance during extreme weather events or abrupt behavioral shifts, where the effective topology of inter-regional dependencies changes more fundamentally. To overcome this limitation, future work could investigate explicitly dynamic graph representations, where both edge weights and connectivity evolve jointly with temporal context. Promising directions include Temporal Graph Networks (TGN), Dynamic Graph Neural Networks (DGNNs), or hybrid approaches combining attention-based adaptivity with data-driven graph updates. Such approaches would enable the model to capture long-term structural drifts while retaining the interpretability and stability of attention mechanisms.

Another limitation concerns the deterministic nature of most graph-based models, which provides point forecasts but no explicit quantification of predictive uncertainty. For operational forecasting, confidence estimates are crucial to support risk-aware decision-making and to ensure transparency for system operators. While ensemble aggregation partially captures epistemic uncertainty, this remains an indirect approach. A natural next step is therefore to develop probabilistic GNNs, e.g. through Bayesian message passing, graph-based quantile regression, or stochastic diffusion mechanisms. Such models could propagate uncertainty across nodes and graph layers, thereby enhancing the interpretability and trustworthiness of graph-based forecasters in operational environments.

Future work will focus on explicitly integrating temporal dependencies within the graph formulation, as well as developing adaptive or season-specific models to better capture structural changes in load behavior. More broadly, extending this framework to multimodal energy data and investigating its scalability to higher-resolution grids constitute promising research directions toward transparent, probabilistic, and reliable forecasting systems.

\paragraph{Disclosure of Interests}%
The authors have no competing interests to declare that are relevant to the content of this article.



\bibliography{refs}

\end{document}